%% file: main.tex
\documentclass{article} 
\usepackage{iclr2022_conference,times}

\usepackage{times}  
\usepackage{helvet}  
\usepackage{courier}  
\usepackage[hyphens]{url}  
\usepackage{graphicx} 
\usepackage{natbib}  
\usepackage{caption} 
\usepackage{bibentry}
\newtheorem{definition}{Definition}

\newtheorem{example}{Example}

\usepackage[noend]{algorithmic}
\usepackage[english]{babel}
\usepackage{amsmath} 
\usepackage{newfloat}
\usepackage{listings}
\usepackage{amssymb}
\usepackage{wrapfig}
\usepackage[ruled,vlined]{algorithm2e}

\input{math_commands.tex}

\usepackage{hyperref}
\usepackage{url}

\usepackage{subcaption}
\title{Learning Dynamic Abstract Representations for Sample-Efficient Reinforcement Learning}

\author{Mehdi Dadvar\\
Arizona State University\\
Tempe, AZ, 85281, USA \\
\texttt{mdadvar@asu.edu} \\
\And
Rashmeet Kaur Nayyar \\
Arizona State University\\
Tempe, AZ, 85281, USA \\
\texttt{rmnayyar@asu.edu} \\
\And
Siddharth Srivastava \\
Arizona State University\\
Tempe, AZ, 85281, USA \\
\texttt{siddharths@asu.edu} \\
}

\newcommand{\alg}{DAR+RL}

\iclrfinaltrue 
\begin{document}

\maketitle

\begin{abstract}
In many real-world problems, the learning agent needs to learn a problem’s abstractions and solution simultaneously. However, most such abstractions need to be designed and refined by hand for different problems and domains of application. This paper presents a novel top-down approach for constructing state abstractions while carrying out reinforcement learning. Starting with state variables and a simulator, it presents a novel domain-independent approach for dynamically computing an abstraction based on the dispersion of Q-values in abstract states as the agent continues acting and learning. Extensive empirical evaluation on multiple domains and problems shows that this approach automatically learns abstractions that are finely-tuned to the problem, yield powerful sample efficiency, and result in the RL agent significantly outperforming existing approaches.
\end{abstract}

\section{Introduction}

It is well known that \emph{good abstract representations} can play a vital role in improving the scalability and efficiency of reinforcement learning (RL)~\citep{sutton2018reinforcement,yu2018towards,konidaris2019necessity}. However, it is not very clear how good abstract representations could be efficiently learned without extensive hand-coding. Several authors have investigated methods for aggregating concrete states based on similarities in value functions but this approach can be difficult to scale as the number of concrete states or the transition graph grows.

This paper presents a novel approach for top-down construction and refinement of abstractions for sample efficient reinforcement learning. Rather than aggregating concrete states based on the agent's experience, our approach starts with a default, auto-generated coarse abstraction that collapses the domain of each state variable (e.g., the location of each taxi and each passenger in the classic taxi world)  to one or two abstract values. This eliminates the need to consider concrete states individually, although this initial abstraction is likely to be too coarse for most practical problems. The overall algorithm proceeds by interleaving the process of refining this abstraction with learning and evaluation of policies, and results in automatically generated, problem and reward-function specific abstractions that aid learning. This process not only helps in creating a succinct representation of cumulative value functions, but it also makes learning  more sample efficient by using the abstraction to locally transfer states' values and  cleaving abstract states only when it is observed that an abstract state contains states featuring a large spread in their value functions. 

\begin{wrapfigure}[15]{r}{0.55\textwidth}
\centering
\includegraphics[width=0.55\columnwidth]{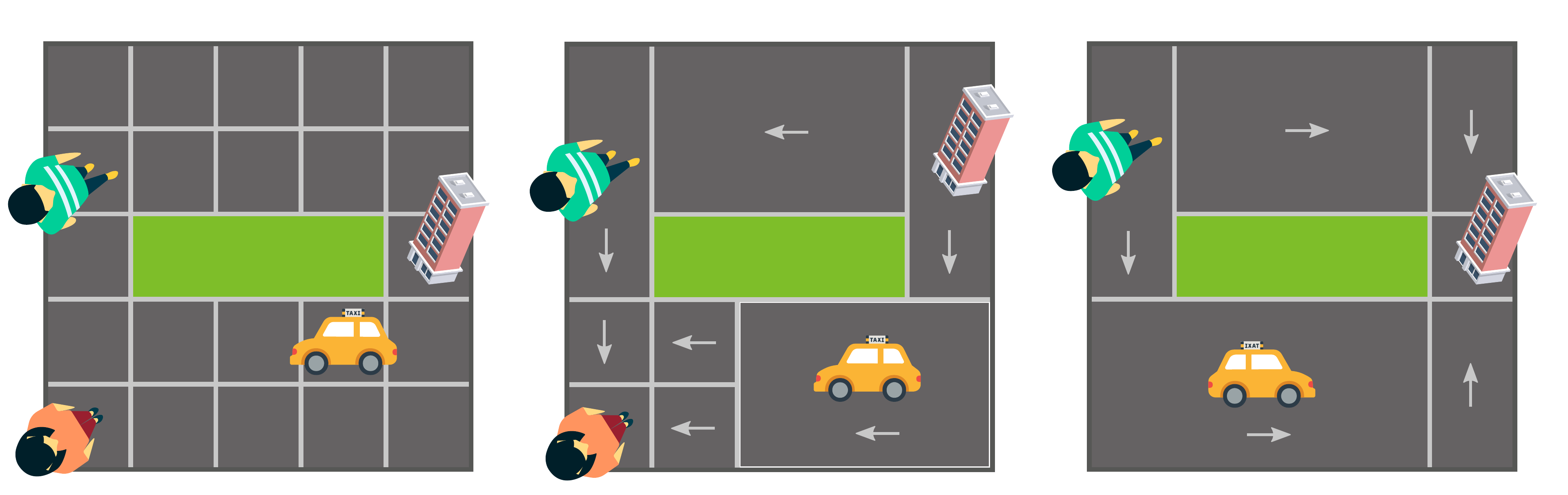} 
\caption{\small Consider a classic taxi world with two passengers and a building as the drop-off location where the green area is impassable (left). Meaningful conditional abstractions can be constructed, for example, for situations where both passengers are at their pickup locations (middle), or one passenger has already been picked-up (right).}
\label{fig:grid-mdp-taxi}
\end{wrapfigure}

This approach is related to research on abstraction for reinforcement learning and on abstraction refinement for model checking~\cite{dams2018abstraction,clarke2000counterexample} (a detailed survey of related work is presented in the next section). However, unlike existing streams of work, we develop a process that automatically generates conditional abstractions, where the final abstraction on the set of values of a variable can depend on the specific values of other variables. For instance, Fig. \ref{fig:grid-mdp-taxi} displays a taxi world where for different values of the state variables (destination and passengers locations), meaningful conditional abstractions are constructed for the taxi location. A meaningful abstraction provides greater details in the taxi-location variable around the passenger location when the taxi needs to pick up a passenger (Fig. \ref{fig:grid-mdp-taxi} (middle)). When the taxi has the passenger, the abstraction should show greater details around the destination (Fig. \ref{fig:grid-mdp-taxi}(right)). Furthermore, our approach goes beyond the concept of counter-example driven abstraction refinement to consider the reward function as well as stochastic dynamics, and it uses measures of dispersion such as the standard deviation of Q-values to drive the refinement process. The main contributions of this paper are mechanisms for building conditional abstraction trees that help compute and represent such abstractions, and the process of interleaving RL episodes with phases of abstraction and refinement. Although this process could be adapted to numerous RL algorithms, we focus on developing and investigating it with Q-learning in this paper.

The presented approach for dynamic abstractions for RL (\alg) can be thought of as a dynamic abstraction scheme because the refinement is tied to the dispersion of Q-values based on the agent's evolving policy during learning. It provides adjustable degrees of compression \citep{abel2016near} where the aggressiveness of abstraction can be controlled by tuning the definition of variation in the dispersion of Q-values. Extensive empirical evaluation on multiple domains and problems shows that this approach automatically learns abstract representations that  effectively draw out similarities across the state space, and yield powerful sample efficiency in learning. Comparative evaluation shows that Q-learning based RL agents enhanced with our approach outperform state-of-the-art RL approaches in both discrete and continuous domains while learning meaningful abstract representations.

The rest of this paper is organized as follows. Sec.\,\ref{sec:related-work} summarizes the related work followed by a discussion on the necessary backgrounds in Sec.\,\ref{sec:back}. Sec.\,\ref{sec:dynamic-abstraction} presents our dynamic abstraction learning method for sample-efficient RL. The empirical evaluations are demonstrated in Sec.\,\ref{sec:emp} followed by the conclusions in Sec.\,\ref{sec:conclusion}.   

\section{Related Work}
\label{sec:related-work}
\textbf{Offline State Abstraction.} Most early studies focus on action-specific \citep{dietterich1999state} and option-specific \citep{jonsson2000automated} state abstraction. Further, \citet{givan2003equivalence} introduced the notion of state equivalence to possibly reduce the state space size by which two states can be aggregated into one abstract state if applying a mutual action leads to equivalence states with similar rewards. Later on, \citet{ravindran2004approximate} relaxed this definition of state equivalence by allowing the actions to be different if there is a valid mapping between them. Offline state abstraction has further been studied for generalization and transfer in RL \citep{karia2022relational} and planning \citep{srivastava2012applicability}.

\textbf{Graph-Theoretic State Abstraction.} \citet{mannor2004dynamic} developed a graph-theoretic state abstraction approach that utilizes the topological similarities of a state transition graph (STG) to aggregate states in an online manner. Mannor's definition of state abstraction follows Givan's notion of equivalence states except they update the partial STG iteratively to find the abstractions. Another comparable method proposed by \citet{chiu2010automatic} carries out spectral graph analysis on STG to decompose the graph into multiple sub-graphs. However, most graph-theoretic analyses on STG, such as computing the eigenvectors in \citeauthor{chiu2010automatic}'s work, can become infeasible for problems with large-scale state space.

\textbf{Monte-Carlo Tree Search (MCTS).} MCTS approaches offer viable and tractable algorithms for large state-space Markovian decision problems \citep{kocsis2006bandit}. \citet{jiang2014improving} demonstrated that proper abstraction effectively enhances the performance of MCTS algorithms. However, their clustering-based state abstraction approach is limited to the states enumerated by their algorithm within the partially expanded tree, which makes it ineffectual when limited samples are available to the planning/learning agent. \citet{anand2015asap} advanced Jiang's method by comprehensively aggregating states and state-action pairs aiming to uncover more symmetries in the domain. Owing to their novel state-action pair abstraction extending Givan and Ravindran's notions of abstractions, \citeauthor{anand2015asap}'s method results in higher quality policies compared to other approaches based on MCTS. However, their bottom-up abstraction scheme makes their method computationally vulnerable to problems with significantly larger state space size. Moreover, their proposed state abstraction method is limited to the explored states since it applies to the partially expanded tree.

\textbf{Counterexample Guided Abstraction Refinement (CEGAR).} CEGAR is a model checking methodology that initially assumes a coarse abstract model and then validates or refines the initial abstraction to eliminate spurious counterexamples to the property that needs to be verified~\citep{clarke2000counterexample}. While most work in this direction focuses on deterministic systems, research on the topic also considers the problem of defining the appropriate notion for a ``counterexample'' in stochastic settings. E.g., \citet{chadha2010counterexample} propose that a counterexample can be considered as a small MDP that violates the desired property. However, searching for such counterexamples can be difficult in the RL setting where the transition function of the MDP is not available. \citet{seipp2018counterexample} developed algorithms for planning in deterministic environments that invoke the CEGAR loop iteratively on the same original task to obtain more efficient abstraction refinement. These methods do not consider the problem of building abstractions for stochastic planning or reinforcement learning.
\vspace{-1em}
\section{Background}
\vspace{-0.5em}
\label{sec:back}
Markov decision Processes (MDPs) \citep{bellman1957markovian, puterman2014markov} are defined as a tuple $\langle \mathcal{S}, \mathcal{A}, \mathcal{T}, \mathcal{R}, \mathcal{\gamma}\rangle$, where $\mathcal{S}$ and $\mathcal{A}$ denote the state and action spaces respectively. Generally, a concrete state $s \in \mathcal{S}$ can be defined as a set of $n$ state variables such that $\mathcal{V} = \{ v_i |  i = 1,...,n \}$. In this paper, we focus on problems where the state is defined using a set of variables. An extension to partially observable settings where the agent receives an image of the state is  a promising direction for future work. $\mathcal{T}: \mathcal{S}\times \mathcal{A} \times \mathcal{S} \rightarrow [0,1]$ is a transition probability function, $\mathcal{R}: \mathcal{S} \times \mathcal{A} \rightarrow \mathbb{R}$ is a reward function, and $\gamma$ is the discount factor. The unknown policy $\pi$ is the solution to an MDP, denoted as $\pi: \mathcal{S} \rightarrow \mathcal{A}$. We consider the RL settings where an agent needs to interact with an environment that can be modeled as an MDP with unknown $\mathcal{T}$. The objective is to learn an optimal policy for this MDP. 

When the size of the space state increases significantly, most of the RL algorithms fail to solve the given MDP due to the \textit{curse of dimensionality}. Abstraction is a dimension reduction mechanism by which the original problem representation maps to a new reduced problem representation \citep{giunchiglia1992theory}. We adopt the general definition of state abstraction proposed by \citet{li2006towards}.

\begin{definition}
\label{def:back}
Let $M = \langle \mathcal{S}, \mathcal{A}, \mathcal{T}, \mathcal{R}, \mathcal{\gamma}\rangle$ be the ground MDP from which the abstract MDP $\bar{M} = \langle \bar{\mathcal{S}}, \mathcal{A}, \bar{\mathcal{T}}, \bar{\mathcal{R}}, \mathcal{\gamma}\rangle$ can be derived via a state abstraction function $\phi: \mathcal{S} \rightarrow \bar{\mathcal{S}}$, where the abstract state mapped to concrete state $s$ is denoted as $\phi(s) \in \bar{\mathcal{S}}$ and $\phi^{-1}(\bar{s})$ is the set of concrete states associated to $\bar{s} \in \bar{S}$. Further, a weighting function over concrete states is denoted as $w(s)$ with $s \in \mathcal{S}$ s.t. for each $\bar{s} \in \bar{\mathcal{S}}$, $\sum_{s \in \phi^{-1}(\bar{s})} w(s) = 1$, where $w(s) \in [0,1]$. Accordingly, the abstract transition probability function $\bar{\mathcal{T}}$ and reward function $\bar{\mathcal{R}}$ are defined as follows:
\begin{align*}
    \bar{\mathcal{R}} (\bar{s},a) &= \sum_{s \in \phi^{-1}(\bar{s})} w(s)\mathcal{R}(s,a), &
\bar{\mathcal{T}}(\bar{s}, a, \bar{s}') &= \sum_{s \in \phi^{-1}(\bar{s})} \sum_{s' \in \phi^{-1}(\bar{s})} w(s)\mathcal{T}(s,a,s').
\end{align*}
\end{definition}

When it comes to the decision-making in an abstract MDP, all concrete states associated with an abstract state $\bar{s} \in \bar{\mathcal{S}}$ are perceived identically. Accordingly, the relation between abstract policy $\pi: \bar{\mathcal{S}} \rightarrow \mathcal{A}$ and the concrete policy $\pi: \mathcal{S} \rightarrow \mathcal{A}$ can be defined as $\pi(s) = \bar{\pi} (\phi(s))$ for all $s \in \mathcal{S}$. Further, the value functions for an abstract MDP are denoted as $V^{\bar{\pi}}(\bar{S})$, $V^*(\bar{S})$, $Q^{\bar{\pi}}(\bar{S},a)$, and $Q^*(\bar{S},a)$. For more on RL and value functions see \cite{sutton2018reinforcement}, for MDPs, see  \citep{bellman1957markovian, puterman2014markov}, and for more on the notion of abstraction, refer \cite{giunchiglia1992theory,li2006towards}.

\section{Our Approach}
\label{sec:dynamic-abstraction}

\subsection{Overview}
Starting with state variables and a simulator, we develop a domain-independent approach for dynamically computing an abstraction based on the dispersion of Q-values in abstract states. The idea of dynamic abstraction is to learn a problem's solution and abstractions simultaneously. We propose a top-down abstraction refinement mechanism by which the learning agent effectively refines an initial coarse abstraction through acting and learning. We illustrate this mechanism with an example.

\begin{wrapfigure}[12]{l}{0.52\textwidth}
\centering
\includegraphics[width=0.52\columnwidth]{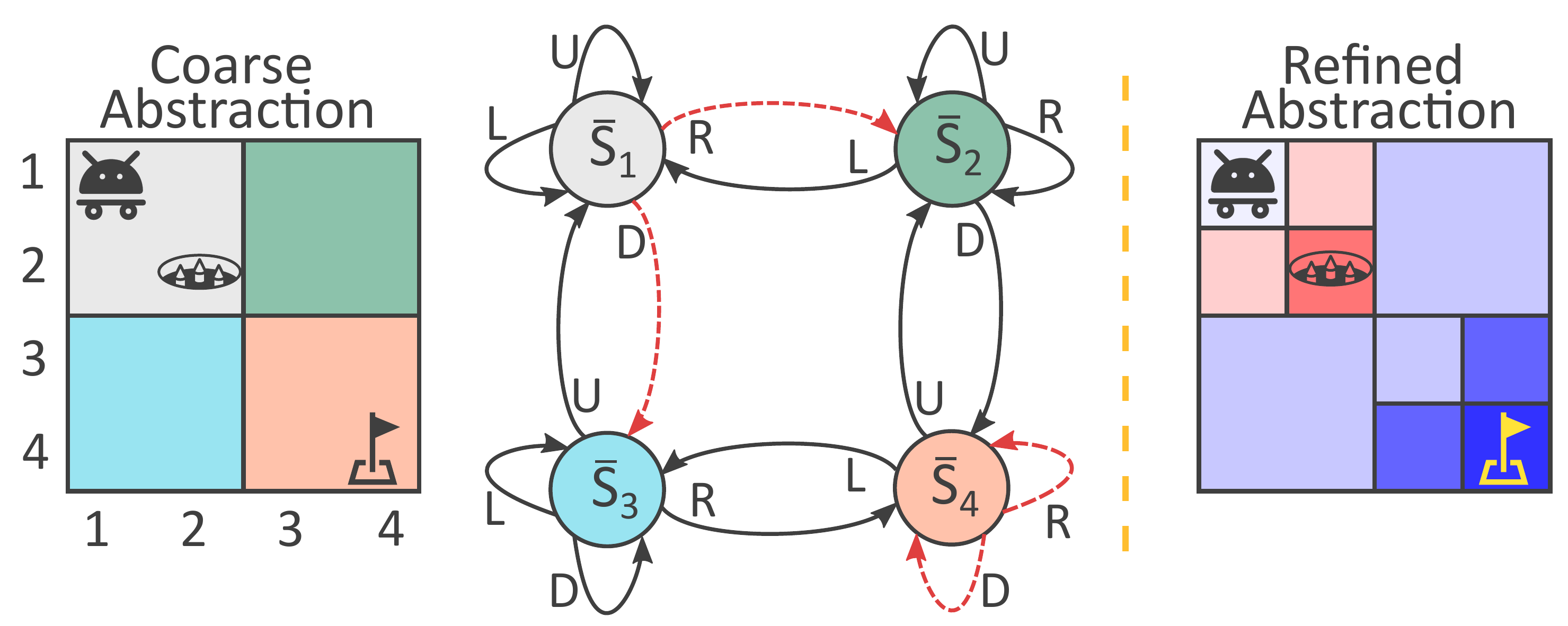}
\caption{\small An example of dynamic heterogeneous abstraction refinement for a Wumpus world.}
\label{fig:grid-mdp}
\end{wrapfigure}

\begin{example}
\label{exmp:grid}
Consider a 4x4 Wumpus world consisting of a pit at (2,2) and a goal at (4,4). In this domain, every movement has a reward -1. Reaching the goal results in a positive reward of 10 and the agents receive a negative reward -10 for falling into the pit. The goal and the pit are the terminal states of the domain. The agent's actions include moving to non-diagonal adjacent cells at each time step s.t. $\mathcal{A}$ = \{up, down, left, right\}.
\end{example}

Considering Example \ref{exmp:grid}, Fig. \ref{fig:grid-mdp} (left) shows a potential initial coarse abstraction in which the domain of each state variable (here horizontal and vertical locations) is split into two abstract values and $\bar{S}_1$ and $\bar{S}_4$ contain the pitfall and goal location respectively. As a result, when learning, the agent will observe a high standard deviation on the values of $Q(\bar{S}_1,right), Q(\bar{S}_1,down), Q(\bar{S}_4,right),$ and $Q(\bar{S}_4,down)$ because of the presence of terminal states with large negative or positive rewards. Guided by this dispersion of Q-values, the initial coarse abstraction should be refined to resolve the observed variations. Fig. \ref{fig:grid-mdp} (right) exemplifies an effective abstraction refinement for Example \ref{exmp:grid} demonstrated as a heatmap of Q-values. Notice that the desired abstraction is a heterogeneous abstraction on the domains of state variable values where the abstraction on a variable depends on the value of the other variables: let $x$ and $y$ be the horizontal and vertical locations of the agent in Example \ref{exmp:grid} respectively and their domain be $\{ 1, 2, 3, 4 \}$. When $y > 2$, the domain of $x$ (originally $\{ 1, 2, 3, 4 \}$) is abstracted into sets $\{ 1,2 \}$, $\{ 3\}$, and $\{ 4 \}$, but when $y \leq 2$, the domain of $x$ is abstracted into sets $\{ 1 \}$, $\{ 2 \}$, and $\{ 3,4 \}$.

The next section (Sec. \ref{sec:hat}) presents our novel approach for automatically computing such abstractions while carrying out RL.

\subsection{Conditional Abstraction Trees}
\label{sec:hat}

The value of a state variable $v_i$ inherently falls within a known range. Partitioning these ranges is one way to construct state abstractions. However, in practice, the abstraction of one state variable is conditioned on a specific range of any other state variables. Accordingly, we need to maintain and update such conditional abstractions via structures that we call Conditional Abstraction Trees (CATs) while constructing the state abstractions. 

\begin{wrapfigure}[19]{l}{0.6\textwidth}
\vspace{-1.8em}
\centering
\includegraphics[width=0.55\columnwidth]{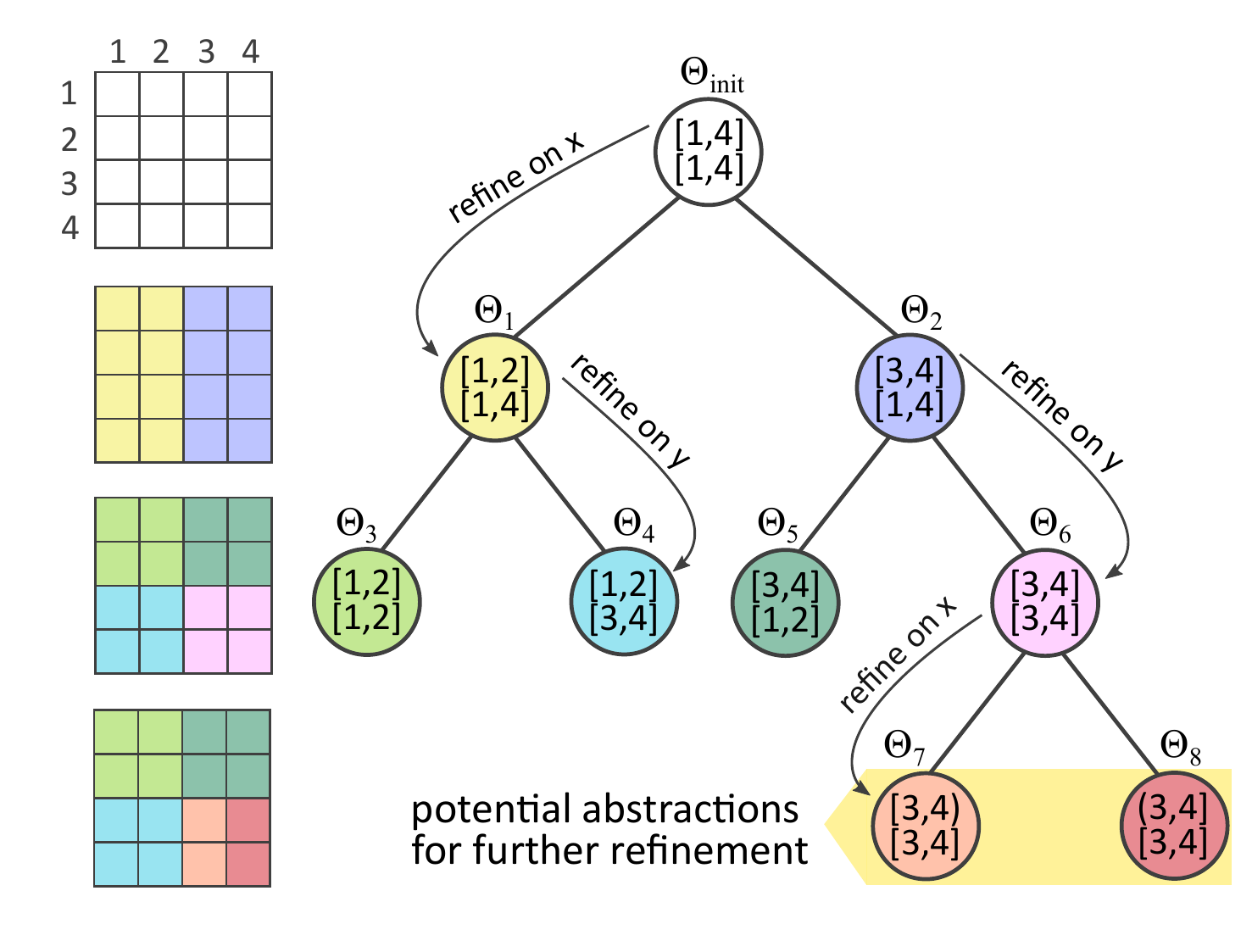} 
\caption{\small This figure illustrates a Conditional Abstraction Tree (CAT) for Example \ref{exmp:grid}. Ranges written inside the nodes represent $\theta_i \in \Theta$. Each node represents a conditional abstraction.}
\label{fig:hat}
\end{wrapfigure}

Fig. \ref{fig:hat} (right) exemplifies a partially expanded CAT for the problem in Example \ref{exmp:grid}. This problem can be represented by two state variables, i.e. agent's horizontal and vertical location denoted as $x$ and $y$ respectively. The tree's root node contains the global ranges (The first range refers to the horizontal location $x$ and the second range refers to the vertical location $y$) for both of these state variables representing an initial coarse abstraction (in white). The annotations visualize how this initial abstraction can be further refined w.r.t a state variable resulting in new conditional abstractions (symmetric annotations are not shown for the sake of readability). The refinement procedure of the Wumpus world associated to each level of the tree is also displayed in Fig. \ref{fig:hat} (left). 

Given the set of state variables $\mathcal{V}$, we define an abstract state using the set of partitions, one for each variable $v_i$, where each partition $\theta_i$ is an interval of the form $[l_i, h_i]$. Thus, a 2-D abstract state for Example \ref{exmp:grid} could be defined by $\theta_1 = [1,4]$ and $\theta_2 = [1,2]$. An abstraction is defined as $\Theta = \{\theta_i | i \in [1,n] \}$, where $n = \lvert \mathcal{V} \rvert$. In fact, CAT is a hierarchical abstraction tree starting with an initial abstraction $\Theta_{init}$ that represents the original range for each state variable $v_i \in s$ s.t. $\Theta_{init} = \{ \theta_i | i \in [1,n] \textrm{ and } l_i = v_i^{min} \textrm{ and } h_i = v_i^{max} \}$, where $v_i^{min}$ and $v_i^{max}$ denote the lower and upper bounds on the range of $v_i$ respectively. In Example \ref{exmp:grid}, there are two state variables so the initial abstraction is $\Theta_{init} = \{ [1,4],[1,4] \}$. The initial abstraction also induces the starting coarse abstraction since the range for each state variable suggests that all values for all state variables is compressed into one abstract state. 

This initial coarse abstraction induced by the initial abstraction $\Theta_{init}$ needs to be further refined so that the learning agent can improve its performance through a more fined representation of the problem. We define a refinement function $\delta(\Theta,i,f)$ that splits the range of partition $\theta_i \in \Theta$ of state variable $v_i$ into $f$ equal ranges resulting in $f$ new abstractions. Now, we formally define the refinement function $\delta(\Theta,i,f)$.

\begin{definition} 
\label{def:refinement}
Let $\Theta = \langle \theta_1, \ldots, \theta_n \rangle$ be an abstract state for a domain with variables $v_1, \ldots v_n$. We define the f-split refinement of $\Theta$ w.r.t. variable $i$ as $\delta(\Theta, i, f) = \{ \Theta^1, \ldots, \Theta^f\}$ where all $\Theta^j$’s are the same as $\Theta$ on $\theta_k$ for $k\ne i$. $\theta_i = [l, h]$ is partitioned with $f$ new boundaries at least $\|\theta\|/f$ values apart: $l, l_1, l_2,\ldots, l_f, h$ where $l_x = l + x\times \lfloor[(h-l)/f]\rfloor$. 
\end{definition}

Next, we need to define the relation between two given abstractions in the form of $\Theta$ in order to determine if one is obtained by refining the other one.    

\begin{definition}
Let $\Psi$ be the set containing all possible abstractions. Given $\Theta_a,\Theta_b \in \Psi$, we say $\Theta_b$ is obtained by refining $\Theta_a$, denoted as $\Theta_b \triangleright \Theta_a$, iff $\; (\forall i \in[1,n]) (\theta_i^b \subseteq \theta_i^a)$. Moreover, $\Theta_b \triangleright \Theta_a \equiv \Theta_a \triangleleft \Theta_b$. Although this definition determines an ancestral relation between $\Theta_a$ and $\Theta_b$, we need to know the factor $f$ by which $\Theta_a$ has been refined to determine if $\Theta_b$ is the direct result of refining $\Theta_a$. We say $\Theta_b$ is obtained directly by refining $\Theta_a$, denoted as $\Theta_b \trianglerighteq \Theta_a$, iff $\; \exists \; i \; (\theta_i^b \subset \theta_i^a)$, $(\forall k_{\neq i} \in[1,n]) (\theta_k^b = \theta_k^a)$ and $\lvert \theta_i^b \rvert \times f = \lvert \theta_i^a \rvert$.
\end{definition}

With these definitions in hand, we can now formally define CAT 
as an undirected tree to construct and maintain the hierarchy of the conditional partitions. Conditional Abstraction Tree (CAT), denoted as $\xi$, represents a tree structure specifying the topology between conditional abstractions in the form of $\Theta$. 

\begin{definition}
\label{def:hat}
A conditional abstraction tree (CAT) is defined as $\xi = \{N, E\}$, where $N$ is the set of nodes and $E$ is the set of edges. Each node in $N$ corresponds to an abstraction $\Theta$, s.t. $N = \{ \Theta_m | m \in [1,n_{\xi}] \}$, where $n_{\xi}$ is the cardinality of CAT, and $\Theta_{root} = \Theta_{init}$, where $\Theta_{root}$ is the root node of the tree. Every parent $\Theta_p$ and child $\Theta_c$ nodes in $\xi$ are connected via an edge $e_p^c$ s.t. $e_p^c \implies \Theta_c \trianglerighteq \Theta_p$. Additionally, $L_{\xi} = \{ \Theta_m | (\forall k \in [1,n_{\xi}]) ( \Theta_m \ntrianglerighteq \Theta_k) \}$ is defined as the set of leaf nodes.
$L_{\xi}$ 
represents the set of abstract states in $\xi$.
\end{definition}
  
\begin{wrapfigure}[13]{r}{0.45\textwidth}

\begin{algorithm}[H]
\caption{State Abstraction}
\label{alg:hat-search}
\textbf{FindAbstract} (CAT $\xi$, $\Theta_{start}$, $s$):
\vspace{-0.9em}
\begin{algorithmic}[1] 
\IF {$(\forall v_i \in s) (v_i \in \theta_i^{start})$} \label{l:inclusion}
\IF {$\Theta_{start} \in L_{\xi}$ }
\STATE \textbf{return} $\Theta_{start}$
\ELSE
\STATE $children \leftarrow Children(\Theta_{start})$
\FOR{$\Theta_{child} \in children$}
\IF {$(\forall v_i \in s) (v_i \in \theta_i^{child})$} \label{l:inclusion_children}
\STATE \textbf{FindAbstract} ($\xi$, $\Theta_{child}$, $s$) \label{l:recursion}
\ENDIF
\ENDFOR
\ENDIF
\ENDIF
\end{algorithmic}
\end{algorithm}
\end{wrapfigure}

Given CAT $\xi$ and the value of a concrete state $s$, the mapping $\phi(s): \mathcal{S} \rightarrow \bar{\mathcal{S}}$ can be done via a level-order tree search starting from $\Theta_{root}$. The corresponding abstract state $\bar{s}$ is in the node $\Theta_{found}$ \textit{iff} $\forall i \in [1,n] \; v_i \in \theta_i^{found}$ (inclusion condition) and $\Theta_{found}$ is a leaf node, i.e. $\Theta_{found} \in L$. Alg. \ref{alg:hat-search} computes the $\phi: \mathcal{S} \rightarrow \bar{\mathcal{S}}$ mapping for a given concrete state $s$ under CAT $\xi$, starting from CAT's root node $\Theta_{root}$. $\texttt{FindAbstract}(\xi, \Theta_{start}, s)$ starts the level-order search from $\Theta_{start}$ and it always finds the corresponding abstract state when $\Theta_{start} = \Theta_{root}$. This algorithm checks the inclusion condition first for $\Theta_{start}$ (Line \ref{l:inclusion} in Alg. \ref{alg:hat-search}). If $\Theta_{Start}$ is not a leaf node, the algorithm checks the inclusion condition for children of $\Theta_{start}$ (Line \ref{l:inclusion_children} in Alg. \ref{alg:hat-search}) and if a child satisfies the condition, $\texttt{FindAbstract}$ gets invoked recursively (Line \ref{l:recursion} in Alg. \ref{alg:hat-search}).

Any state abstraction under a given CAT $\xi$ induces an abstract representation of the underlying concrete MDP $M$. Thus, an MDP $M$ can have two abstract representations $\bar{M}_a$ and $\bar{M}_b$ under two CATs $\xi_a$ and $\xi_b$ respectively. We define a relational operation to decide which abstract MDP is finer.

\begin{definition}
Given MDPs $\bar{M}_a$ and $\bar{M}_b$ abstracted under $\xi_a$ and $\xi_b$, we say $\bar{M}_a$ is strictly finer than $\bar{M}_b$, denoted as $\bar{m}_a \succ \bar{m}_b$, iff $\forall \Theta^a \in L_{\xi_a} \; \exists \Theta^b \in L_{\xi_b} \; (\Theta^a \trianglerighteq \Theta^b)$. We also say $\bar{M}_a$ is finer than $\bar{M}_b$, denoted as $\bar{M}_a \succeq \bar{M}_a$, iff $\forall \Theta^a \in L_{\xi_a} \; \exists \Theta^b \in L_{\xi_b} \; (\Theta^a \trianglerighteq \Theta^b \vee \Theta^a = \Theta^b)$.   
\end{definition}

\subsection {Learning Dynamic Abstractions}
\label{sec:learning-dynamic-abs}
Definition \ref{def:hat} formalizes the abstraction tree by which the mapping $\phi(s): \mathcal{S} \rightarrow \bar{\mathcal{S}}$ can be performed using a level-order search (see Alg. \ref{alg:hat-search}), while Definition \ref{def:refinement} explains how a node of a CAT can be refined with respect to a state variable $v_i$ through the refinement function $\delta(\Theta, i, f)$. However, our objective is to interleave RL episodes with phases of abstraction refinement leading to an enhanced abstract policy $\bar{\pi}$ for a given concrete MDP $M$. We need to develop a mechanism that 1) observes the dispersion of Q-values while the RL agent is acting and learning through the abstract MDP $\bar{M}$, and 2) refines unstable abstract states w.r.t a state variable. 

Therefore, our approach, Dynamic Abstractions for RL (\alg{}), consists of three phases: 1) the RL agent performs Q-learning over the abstract state space $\bar{\mathcal{S}}$ defined by leaves of the current CAT and learns an abstract policy $\bar{\pi}$; 2) the RL agent continues interacting with the environment via the abstract policy $\bar{\pi}$ and \alg{} evaluates the computed abstraction by observing the dispersion of Q-values; and 3) \alg{} refines the current abstraction by finding unstable abstract states in $\xi$. One needs to blame a state variable $v_i$ for each unstable state since the refinement can be conducted w.r.t to one state variable as defined in Definition \ref{def:refinement}.

Let $\beta(M, \xi, \bar{\pi})$ denote the evaluation function which is simply an RL routine where the learning agent interacts with an MDP $M$ through a fixed policy $\bar{\pi}$ under the abstraction computed by $\xi$ for one single episode; Throughout one episode of evaluation, the observed dispersion of Q-values is defined as $D = \{ d_m | m \in [1, n_{step}] \}$. The observed dispersion $D$ is the set of observed $Q^{\bar{\pi}}(\bar{S},a)$ values for one episode (up to $n_{step}$ steps) of evaluation.

\begin{wrapfigure}[23]{l}{0.52\textwidth}
\begin{algorithm}[H]
\caption{Learning Dynamic Abstractions}
\label{alg:main}
\textbf{Input}: $M, f$ \\
\textbf{Output}: $\bar{M}, \xi, \bar{\pi}$
\begin{algorithmic}[1] 
\STATE initialize $\Theta_{init}$, $\xi$, and $\bar{Q}$ \label{line:initialize}
\FOR{$episode = 1, n_{epi}$} \label{line:routine_begin}
\STATE $s \leftarrow \texttt{reset()}$
\FOR {$steps$ in $episode$}
\STATE $\bar{s} \leftarrow \texttt{FindAbstract}(\xi, \Theta_{init}, s)$
\STATE $a \leftarrow \bar{\pi}(\bar{s})$
\STATE $s', \bar{r}, done  \leftarrow \texttt{step}(\texttt{extend}(a))$ \label{line:extend}
\STATE $\bar{s'} \leftarrow \texttt{FindAbstract}(\xi, \Theta_{init}, s')$
\STATE $\bar{\pi} \leftarrow \texttt{train}^{\bar{\pi}}(\bar{s}, \bar{s}', a, \bar{r})$
\STATE $s, \bar{s} \leftarrow s', \bar{s}'$ \label{line:routine-end}
\ENDFOR
\IF {$\bar{M}$ needs refinement} \label{line:refinement-condition}
\FOR {$e = 1, n_{eval}$} \label{line:start-eval}
\STATE $\Gamma.\texttt{append} (\texttt{evaluate} (M, \xi, \bar{\pi}))$ \label{line:sim}
\ENDFOR
\STATE $unstable \leftarrow \texttt{UnstableState}(\Gamma)$ \label{line:unstable}
\FOR {each $\Theta$ in $unstable$} \label{line:update-tree-loop}
\STATE $i \leftarrow \texttt{UnstableVar}(\Gamma, \Theta)$
\STATE $nodes \leftarrow \texttt{refine}(\Theta, i, f)$
\STATE $\xi \leftarrow \texttt{UpdateTree}(\xi, \Theta, nodes)$ \label{line:end-update-tree}
\ENDFOR
\ENDIF
\ENDFOR
\STATE \textbf{return} $\bar{M}, \xi, \bar{\pi}$
\end{algorithmic}
\end{algorithm}
\end{wrapfigure}
\vspace{-0.5em}

To obtain a better exploration over the abstract states, the evaluation function $\beta(M, \xi, \bar{\pi})$ needs to be executed for $n_{eval} > 1$ episodes. That being said, $\Gamma$ denotes all observed dispersion obtained by executing the evaluation function for $n_{eval}$ episodes, where  $\Gamma = \{ D_m | m \in [1,n_{eval}] \}$. Let $\texttt{UnstableState}(\Gamma)$ denote a function that finds the set of unstable states in the form of $\Theta$ based on the dispersion of Q-values in $\Gamma$. Besides, $\texttt{UnstableVar}(\Gamma, \Theta)$ denotes a function that finds the accountable state variable for each unstable state, given the dispersion log $\Gamma$. Altogether, within the \alg{} learning, evaluation, and refinement phases, the learning agent learns the solution to the MDP $M$ while learning the dynamic abstractions. 

\subsection{\alg{} Algorithm}
Alg. \ref{alg:main} illustrates the procedure by which the agent learns an MDP's solution and abstractions simultaneously through learning, evaluation, and refinement phases explained in Sec. \ref{sec:learning-dynamic-abs}. First, the initial coarse abstraction needs to be automatically constructed through initializing $\Theta_{init}$, based on the known ranges for each state variable $v_i$, and constructing a CAT $\xi$ with only the root node (Line \ref{line:initialize} in Alg. \ref{alg:main}). The initial $\xi$ induces an abstract MDP $\bar{M}$ for the given MDP $M$. Then, the learning phase of \alg{} starts by employing the Q-learning routine (Lines \ref{line:routine_begin} to \ref{line:routine-end} in Alg. \ref{alg:main}). In other words, throughout the learning phase (lines \ref{line:routine_begin} to \ref{line:routine_begin}) Alg. \ref{alg:main} implement the vanilla Q-learning over abstract states computed from the CAT. In this phase, induced by the computed state abstraction, extended actions (taking a concrete action repeatedly until the agent reaches a new abstract state, blockage, or a terminal concrete state) are applied to the environment instead of the concrete actions (Line \ref{line:extend} in Alg. \ref{alg:main}). As the result, the agent enhances the abstract policy $\bar{\pi}$ to learn the solution to the abstract MDP $\bar{M}$. 

Since the initial abstraction is likely to be too coarse, \alg{} checks the refinement condition (Line \ref{line:refinement-condition} in Alg. \ref{alg:main}) at the end of each learning episode to initiate an evaluation phase followed by a refinement phase. We set \alg{} to check the recent success rate of the RL agent every $n_{check}$ episodes where the refinement condition evaluate to true if the success rate is below some threshold. The choice of the refinement condition introduces a trade-off. On one hand, we want to obtain a near optimal abstraction that enables the agent to learn the solution effectively. On the other hand, the abstract policy $\bar{\pi}$ should be trained enough to be used in the evaluation phase for refinement purposes. When the refinement condition is true, the algorithm runs the evaluation function $\beta$ (a standard Q-learning routine with a fixed policy) for $n_{eval}$ episodes (Line \ref{line:sim} in Alg. \ref{alg:main}). During the evaluation phase, it is likely to encounter an abstract state $\bar{s}$ multiple times. Since the policy is fixed in this phase, comparing different Q-values for the same pair of state-action ($\bar{s}, \bar{\pi}(a)$) exposes potential inconsistencies in the abstract state $\bar{s}$. To capture such inconsistencies, for all observed abstract states, \alg{} logs different computed Q-values and store them in $\Gamma$. Then, for each abstract state $\bar{s}$ in $\Gamma$, \alg{} calculates the normalized standard deviation of all logged $Q(\bar{s}, \bar{\pi}(a))$ (Line \ref{line:unstable} in Alg \ref{alg:main}). Considering these calculated values, $\texttt{UnstabelState}(\Gamma)$ finds the unstable states (abstract states with high variations) using clustering techniques. Finally, the unstable states are refined and the abstraction tree is updated accordingly in Lines \ref{line:update-tree-loop} to \ref{line:end-update-tree}.

\section{Empirical Evaluations}
\label{sec:emp}
To assess the performance of \alg, we conducted empirical analysis on three discrete domains: Office World adapted from \cite{icarte2018using}, Wumpus World derived from \cite{stuart2010artificial}, Taxi World adapted from the OpenAI Gym environment Taxi-v3 (\url{https://www.gymlibrary.ml/environments/toy_text/taxi/}) and introduced by \cite{dietterich2000hierarchical}, and one continuous domain: Water World based on \cite{karpathyreinforcejs,icarte2018using}. All of the domains used are stochastic continuous/discrete problems with varying dimensionality (from 2 to 14). All the details regarding the domains and task descriptions are included in the supplementary document. We aim to investigate the following:
\begin{itemize}
    \item Does \alg{} improve the sample efficiency of vanilla Q-learning without any expert knowledge?
    \item Is \alg{} scalable to high-dimensional tasks?
    \item Does \alg{} recognize similar abstractions for similar sub-problems in a larger task?
\end{itemize}

For the comparative study, we selected the following baselines: (1) Option-critic \cite{bacon2017option}, (2) JIRP \cite{xu2020joint}, (3) tabular Q-learning \cite{watkins1992q}, (4) DQN \cite{mnih2013playing}, (5) A2C \cite{mnih2016asynchronous}, and (5) PPO \cite{schulman2017proximal}. 
Option-critic is a Hierarchical RL (HRL) approach that discovers options autonomously while learning option policies simultaneously. JIRP automatically infers reward machines and policies for RL. We chose these state-of-the-art methods as baselines as they do not require expert knowledge as input. 
We also compared with deep RL methods: DQN, A2C, and PPO. The details of parameters and hyperparameters are included in the supplementary document.

For each domain, we executed 10 independent runs for each algorithm and report the mean success rates averaged over last 100 training episodes along with the standard deviations. We also report normalized cumulative reward for each domain and method obtained by evaluating the agent on 10 simulation runs, after stopping training at intervals of 10 episodes. We use implementation of DQN, A2C, and PPO from the Stable-Baselines3 by \cite{raffin2019stable}. Our code is included in the supplementary material. We now discuss our results and analysis in detail below.

\begin{figure}[t]
     \centering
         \centering
         \includegraphics[width=\textwidth]{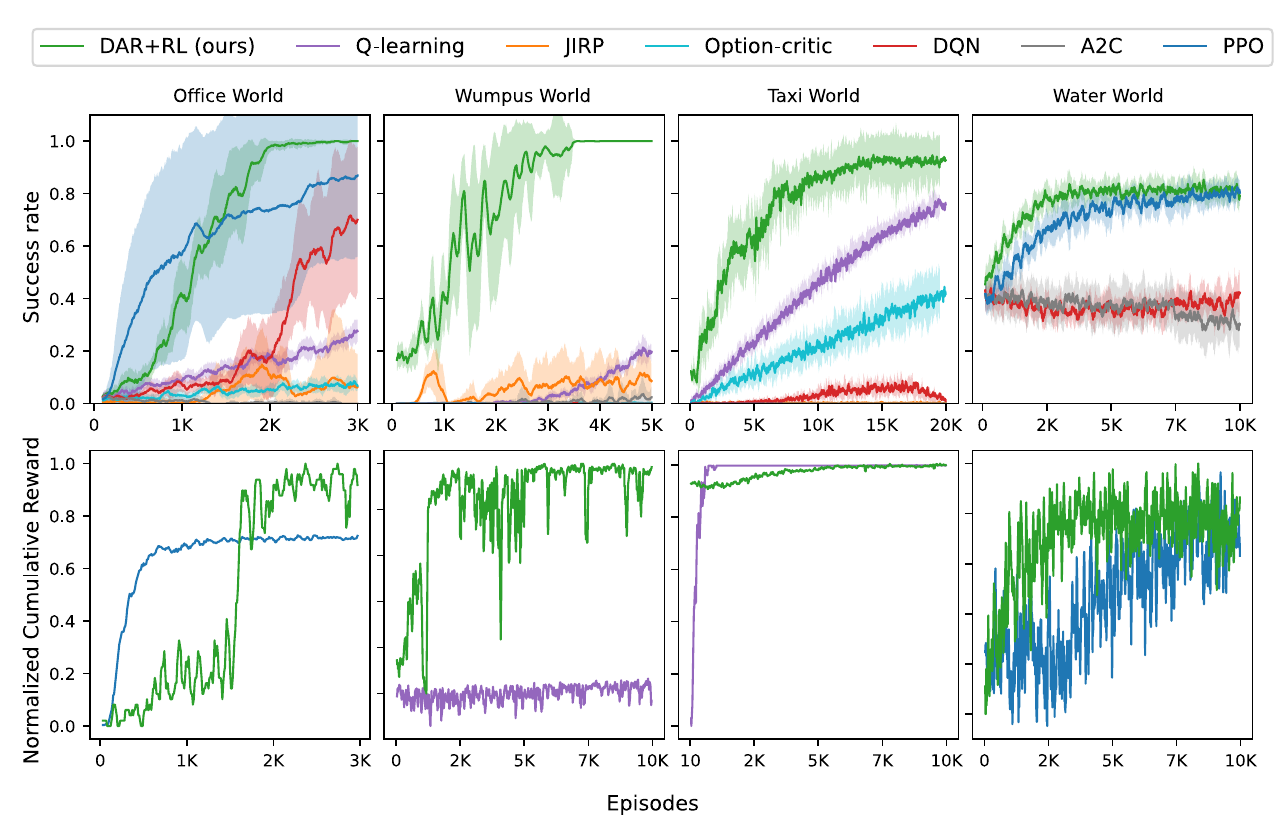}
        \caption{\small (Top) Success rates (mean and standard deviation) for 10 independent runs averaged over last 100 training episodes for all the methods, and (Bottom) normalized cumulative reward for 10 simulation runs obtained every 10 training episodes for DAR+RL (ours) and the second-best performing baseline for Office World (36x36), Wumpus World (64x64), Taxi World (30x30), and Water World (300x300) domains.
        }
        \label{fig:result}
\end{figure}

\subsection{Results}
\label{sec:results}
Fig. \ref{fig:result} (Top) shows comparison among success rates achieved by all the methods on all the domains. In Office world, \alg{} outperforms all the baselines and almost converges to a success rate of 1 in around 2000 episodes, whereas, PPO reaches an approximate success rate of 0.8 in 2500 episodes and has a high standard deviation. DQN reaches a success rate of only 0.65 within 3000 episodes and rest of the baselines struggle to learn and are stuck below 0.2 success rate. In Wumpus world, \alg{} converges to success rate of 1 within 4000 episodes and significantly outperform all the baselines which are stuck below a success rate of only 0.2. In Taxi world, \alg{} achieves a success rate of almost 1.0 within 12000 episodes of training, while Q-learning and Option-critic perform better than other baselines achieving approximate success rates of 0.75 and 0.4 respectively within 20000 episodes. Even in the continuous Water world domain, \alg{} learns slightly faster than PPO while all other baselines perform poorly and are stuck below a success rate of 0.4. We performed further evaluation on \alg{} and the second-best performing baseline on each domain as shown in the Fig. \ref{fig:result} (Bottom) by evaluating the policies learned by the agent. In Office and Water worlds, \alg{} outperforms PPO, and in Wumpus world, \alg{} gains significantly higher cumulative reward than Q-learning. 

\subsection{Analysis}

\textbf{Sample efficiency in the absence of input expert knowledge.}
The results presented in Section \ref{sec:results} demonstrate that \alg{}'s performance is superior to all baselines in both discrete and continuous domains. This is categorically the effect of the learned conditional abstractions by \alg{} made available to the vanilla Q-learning algorithm. This effect can be perceived from two perspectives: 1) the meaningful conditional abstractions that are automatically constructed by \alg{} spotlight the most informative aspects of the state space, leading to more sample-efficient learning; and 2) the Q-learning agent benefits from significantly higher levels of exploration over state and action spaces due to the nature of abstraction. This intense exploration can cause more penalization of the agent at the early stages of learning (see cumulative rewards of \alg{} in taxi and office world) but eventually leads to faster learning and superior performance reflected in the success rate. 

\textbf{Scalability to high-dimensional tasks.}
RL algorithms that learn policy $\pi$ from a concrete MDP $M$ suffer from the curse of dimensionality as the size of the state space increases. This explains why most of the baselines fail to learn the Wumpus world, as a basic domain, when the size of the grid increases drastically, as shown in Fig. \ref{fig:result}. In contrast, the top-down abstraction refinement scheme of \alg{} scales effectively to problems with relatively larger state space. As a result, the abstract representations learned by \alg{} empowered vanilla Q-Learning algorithm to learn those problems relatively fast and efficiently. We conducted further experiments on scalability and computational complexity of \alg{} and baselines and the results are presented in the supplementary document.

\begin{wrapfigure}[11]{r}{0.4\textwidth}
\centering
\includegraphics[width=0.4\columnwidth]{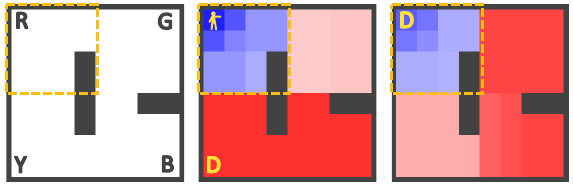} 
\caption{\small Drawing out similarities across state space of a 8$\times$8 taxi world via \alg's automatic abstraction.}
\label{fig:heat}
\end{wrapfigure}

\textbf{Abstractions learned in similar sub-problems.}
One important property of \alg's framework is to construct identical abstractions across the state space for similar sub-problems. This capability of \alg{} is critically beneficial in large problems where options can be generalized across identically constructed abstractions. Fig. \ref{fig:heat} demonstrates two constructed conditional abstractions by \alg{} for an 8$\times$8 taxi world. In Fig. \ref{fig:heat} (middle), the passenger is located at top-left and the destination is located at the bottom-left of the map. Besides, in Fig. \ref{fig:heat} (right), the passenger is in the taxi and the destination is located at the top-left. In both cases, the agent should reach  the top-left cell of the map which implies a similarity. \alg{} discovered this similarity automatically as seen from the generated identical abstractions (highlighted area) for both the cases.   

\section{Conclusion}
\label{sec:conclusion}
\vspace{-0.9em}
We presented a novel approach (\alg{}) for simultaneously learning dynamic abstract representations along with the solution to problems formulated as an MDP. The overall algorithm of \alg{} proceeds by interleaving the process of refining a coarse initial abstraction with learning and evaluation of policies for the underlying RL agent (Q-learning). Besides, we introduced conditional abstraction trees to compute and represent such refined abstractions throughout the \alg{} procedure. Extensive empirical evaluation on multiple domains of problems demonstrated that \alg{} effectively enables vanilla Q-learning algorithm to learn the solution to large discrete and continuous problems, with dynamic representations, where state-of-the-art RL algorithms are outperformed. This superior performance of vanilla Q-learning compared to algorithms with complex neural-network-based architecture such as PPO and A2C is due to \alg{}'s scalable abstraction construction scheme that effectively draws out similarities across the state space and yields powerful sample efficiency in learning. Future work will consider automatic discovery of generalizable options utilizing the constructed conditional abstract representations by \alg{}. 



\bibliography{iclr2022_conference}
\bibliographystyle{iclr2022_conference}

\appendix

\section{Scalability Study}

We conducted experiments (see Fig. \ref{fig:result}) to test the scalability of DAR+RL, Q-learning, and PPO on Office World problems with increasing complexity. 
It shows that DAR+RL has greater scalability than the baselines. The results also indicate that (a) DRL methods do better on smaller problem instances that are less “complex” and are unable to handle increasing complexity and (b) methods designed for image-based RL do not directly scale in RL problems such as those used in this work. Thus, DAR+RL addresses the challenges with scalability of RL to tasks whose states cannot be easily expressed as images or robot configuration states.

\begin{figure}[h]
     \centering
         \centering
         \includegraphics[width=\textwidth]{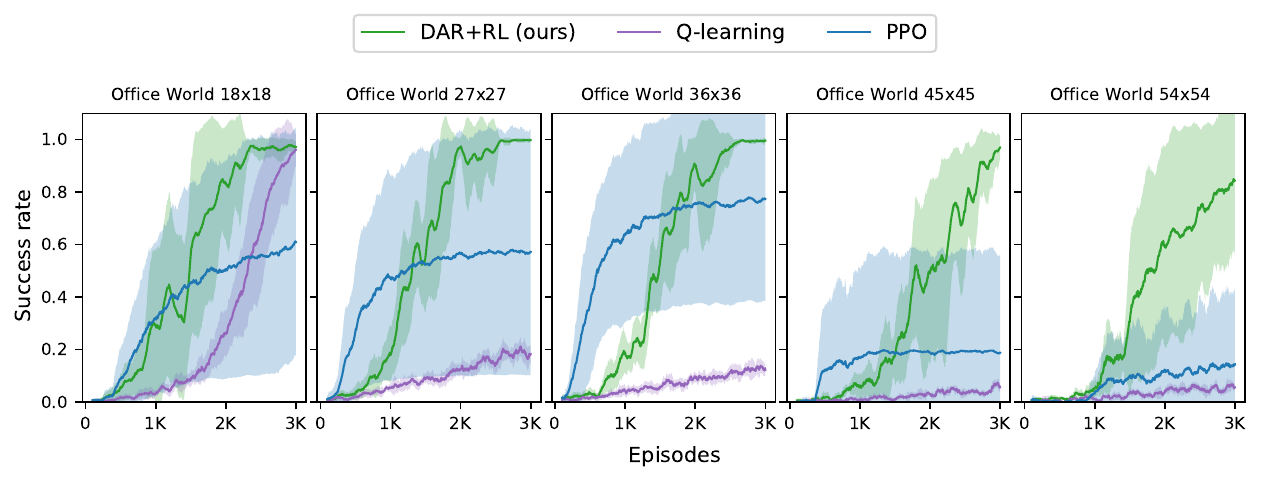}
        \caption{\small Scalability of DAR+RL (our method), Q-learning, and PPO on Office World problems with increasing complexity i.e. increasing ranges of state variables. The title refers to the problem size, y-axis shows average success rates and standard deviations for 10 independent runs averaged over last 100 training episodes, and x-axis shows episodes. The maximum episode lengths used for Office World problems with size 18x18, 27x27, 36x36, 45x45, and 54x54 are 250, 500, 700, 1000, and 1500 respectively.
        }
        \label{fig:result}
\end{figure}

\begin{figure}[h]
     \centering
         \centering
         \includegraphics[width=\textwidth]{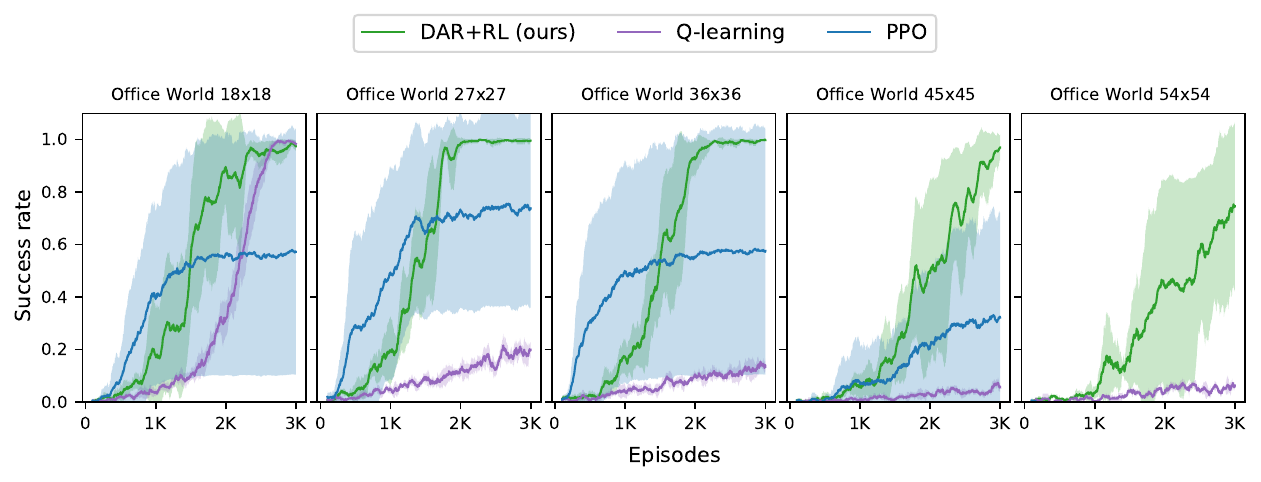}
        \caption{\small Scalability of DAR+RL (our method), Q-learning, and PPO with small neural network. This is a replication of the scalability study reported in Fig. \ref{fig:result} except we ran PPO with a smaller neural network architecture (two hidden layers with 16 neurons per hidden layer).
        }
        \label{fig:result2}
\end{figure}

We replicated the scalability study with an exact condition compared to Fig. \ref{fig:result} except we altered the neural network architecture of PPO to study the effect of the neural network architecture of deep RL algorithms on their scalability, as shown in Fig. \ref{fig:result2}. To this end, we reduced the size of PPO's network architecture from 64 to 16 neurons per hidden layer, where two hidden layers were utilized in both cases. The results indicate that reducing the network size does not improve the performance of PPO and rejects the hypothesis that the original architecture used in the paper for deep RL baselines might be over parameterized or excessively large for the given test problems.

\section{Time Complexity Analysis}
The worst case of computational complexity of the learning and evaluation phases of DAR+RL is similar to that of the underlying RL algorithm that it is used (Q-learning). The refinement phase consists of a CAT search for an unstable state with a time complexity of $O(n\log n)$ and a split operation which is linear in the number of state variables. 

Tab. \ref{tab:my_label} shows the time (mean and standard deviation computed for 10 runs) taken for DAR+RL, Q-learning, and PPO for solving Office World problems with increasing problem complexity. All experiments were executed on 5.0 GHz Intel i9 CPUs with 64 GB RAM running Ubuntu 18.04.

\begin{table}[h!]
    \centering
    \begin{tabular}{|p{25mm}|p{30mm}|p{30mm}|p{30mm}|}
        \hline
         Office World $\newline$ problem size & Time (s) $\pm$ std dev $\newline$ by DAR+RL & Time (s) $\pm$ std dev $\newline$ by Q-learning & Time (s) $\pm$ std dev $\newline$ by PPO\\
          \hline
         18x18 & $302.75 \pm 29.0$ & $97.69 \pm 4.7$ & $2843.42 \pm 959.17$ \\
         27x27 & $391.36 \pm 28.5 $ & $441.8 \pm 13.85$ & $ 4956.8 \pm 2458.74$ \\
         36x36 & $535.71 \pm 54.6$ & $1174.84 \pm 46.23$ &  $8428.24 \pm 2867.52$ \\
         45x45 & $416.41 \pm 58.94$ & $1322.26 \pm 45.87$ & $11463.28 \pm 3309.09$\\
         54x54 & $1010.52 \pm 219.98$ & $7750.53 \pm 308.79$ & $15293.57 \pm 5815.63$\\
         \hline
    \end{tabular}
    \caption{Total time taken (mean and standard deviation) by DAR+RL, Q-learning, and PPO to solve Office World problems with increasing complexity.}
    \label{tab:my_label}
\end{table}

We found DAR+RL to be surprisingly efficient in terms of runtime. Although Q-learning completed before our approach for small problems, DAR+RL is significantly faster than Q-learning when the problem size increases even when the time for abstraction refinement is taken into account. The reason for this performance boost is that, in practice, DAR+RL performs significantly fewer computations than it would require to solve the underlying MDP due to the abstraction that it builds on the fly. Although the abstract MDP becomes finer after each refinement phase, the state space size of this abstract MDP is still significantly smaller than the concrete MDP.

\section{Hyperparameters}
We used standard architectures for A2C, PPO, DQN from StableBaselines3 (https://github.com/DLR-RM/stable-baselines3) and Option-Critic (https://github.com/lweitkamp/option-critic-pytorch). We use the open-source code available for the state-of-the-art baseline JIRP (https://github.com/logic-and-learning/AdvisoRL).

\alg{}'s parameters are $n_{check}$ and the threshold value $t_{succ}$ for the refinement condition, the cap $k$ for the maximum number of unstable states that can be refined in each refinement phase, and $n_{eval}$ for the duration of the evaluation phase. The same parameter values were used across all our experiments except for cap $k$ which was set proportionally to the size of the problem. In contrast, we had to conduct significant hyper-parameter exploration for the baselines because the default settings led to insignificant learning. 

Tab. \ref{tab:tab1}, \ref{tab:tab2}, \ref{tab:tab3}, \ref{tab:tab4} show the important hyperparameters used for all the domains and methods.

\begin{table}[h!]
    \centering
    \begin{tabular}{|p{37mm}|p{12mm}|p{10mm}|p{9mm}|p{9mm}|p{9mm}|p{9mm}|p{9mm}|}
    \hline
        Hyperparameters & DAR+RL & Q-learning & Option-critic & JIRP & A2C & DQN & PPO \\
        \hline
        Threshold ($t_{succ}$) & 0.8 & $-$ & $-$ & $-$ & $-$ & $-$ & $-$\\
        $n_{check}$ & 100 & $-$ & $-$ & $-$ & $-$ & $-$ & $-$\\
        $n_{eval}$ & 100 & $-$ & $-$& $-$ & $-$& $-$ & $-$\\
        Cap ($k$) & 20 & $-$& $-$& $-$& $-$& $-$& $-$\\
        Exploration rate $(\epsilon)$ & 1.0 & 1.0 & 1.0 & 0.4 & $-$ & 1.0 & $-$ \\
        Minimum exploration rate & 0.05 & 0.05 & 0.05 & 0.05 & $-$ & 0.05 & $-$ \\
        Exploration decay & 0.991 & 0.991 & 0.9991 & 0.9991 & $-$ & $-$ & $-$ \\
        Exploration fraction & $-$ & $-$ & $-$ & $-$ & $-$ & 1.0 & $-$ \\
        Learning rate $(\alpha)$ & 0.05 & 0.05 & 0.05 & 1e-4 & 7e-4 & 2e-4 & 2e-4\\ 
        Discount factor $(\gamma)$ & 0.95 & 0.95 & 0.95 & 0.95 & 0.95 & 0.95 & 0.95 \\
        Number of episodes & 10000 & 10000 & 10000 & 10000 & 10000 & 10000 & 10000 \\
        Maximum episode length & 1200 & 1200 & 1200 & 1200 & 1200 & 1200 & 1200\\
        Options & - & - & 8 & - & - & - & -\\
        \hline
    \end{tabular}
    \caption{Parameters used in Wumpus World.}
    \label{tab:tab1}
\end{table}

\begin{table}[h!]
    \centering
    \begin{tabular}{|p{37mm}|p{12mm}|p{10mm}|p{9mm}|p{9mm}|p{9mm}|p{9mm}|p{9mm}|}
    \hline
        Hyperparameters & DAR+RL & Q-learning & Option-critic & JIRP & A2C & DQN & PPO \\
        \hline
        Threshold ($t_{succ}$) & 0.8 & $-$ & $-$ & $-$ & $-$ & $-$ & $-$\\
        $n_{check}$ & 100 & $-$ & $-$ & $-$ & $-$ & $-$ & $-$\\
        $n_{eval}$ & 100 & $-$ & $-$& $-$ & $-$& $-$ & $-$\\
        Cap ($k$) & 20 & $-$& $-$& $-$& $-$& $-$& $-$\\
        Exploration rate $(\epsilon)$ & 1.0 & 1.0 & 1.0 & 0.4 & $-$ & 1.0 & $-$ \\
        Minimum exploration rate & 0.05 & 0.05 & 0.05 & 0.05 & $-$ & 0.05 & $-$ \\
        Exploration decay & 0.9992 & 0.9992 & 0.9992 & 0.9992 & $-$ & $-$ & $-$ \\
        Exploration fraction & $-$ & $-$ & $-$ & $-$ & $-$ & 1.0 & $-$ \\
        Learning rate $(\alpha)$ & 0.05 & 0.05 & 0.05 & 1e-4 & 8e-4 & 1e-4 & 3e-4\\ 
        Discount factor $(\gamma)$ & 0.99 & 0.99 & 0.99 & 0.99 & 0.99 & 0.99 & 0.99 \\
        Number of episodes & 3000 & 3000 & 3000 & 3000 & 3000 & 3000 & 3000 \\
        Maximum episode length & 1000 & 1000 & 1000 & 1000 & 1000 & 1000 & 1000\\
        Options & - & - & 8 & - & - & - & -\\

        \hline
    \end{tabular}
    \caption{Parameters used in Office World.}
    \label{tab:tab2}
\end{table}

\begin{table}[h!]
    \centering
    \begin{tabular}{|p{37mm}|p{12mm}|p{10mm}|p{9mm}|p{9mm}|p{9mm}|p{9mm}|p{9mm}|}
    \hline
        Hyperparameters & DAR+RL & Q-learning & Option-critic & JIRP & A2C & DQN & PPO \\
        \hline
        Threshold ($t_{succ}$) & 0.8 & $-$ & $-$ & $-$ & $-$ & $-$ & $-$\\
        $n_{check}$ & 100 & $-$ & $-$ & $-$ & $-$ & $-$ & $-$\\
        $n_{eval}$ & 100 & $-$ & $-$& $-$ & $-$& $-$ & $-$\\
        Cap ($k$) & 10 & $-$& $-$& $-$& $-$& $-$& $-$\\
        Exploration rate $(\epsilon)$ & 1.0 & 1.0 & 1.0 & 0.4 & $-$ & 1.0 & $-$ \\
        Minimum exploration rate & 0.05 & 0.05 & 0.05 & 0.05 & $-$ & 0.05 & $-$ \\
        Exploration decay & 0.9992 & 0.9992 & 0.9992 & 0.9992 & $-$ & $-$ & $-$ \\
        Exploration fraction & $-$ & $-$ & $-$ & $-$ & $-$ & 1.0 & $-$ \\
        Learning rate $(\alpha)$ & 0.05 & 0.05 &  0.05 & 5e-5 & 7e-4 & 1e-4 & 2e-4\\ 
        Discount factor $(\gamma)$ & 0.999 & 0.999 & 0.999 & 0.999 & 0.999 & 0.999 & 0.999 \\
        Number of episodes & 20000 & 20000 & 20000 & 20000 & 20000 & 20000 & 20000 \\
        Maximum episode length & 1500 & 1500 & 1500 & 1500 & 1500 & 1500 & 1500\\
        Options & - & - & 8 & - & - & - & -\\
        \hline
    \end{tabular}
    \caption{Parameters used in Taxi World.}
    \label{tab:tab3}
\end{table}

\begin{table}[h!]
    \centering
    \begin{tabular}{|p{37mm}|p{12mm}|p{10mm}|p{9mm}|p{9mm}|p{9mm}|p{9mm}|p{9mm}|}
    \hline
        Hyperparameters & DAR+RL & A2C & DQN & PPO \\
        \hline
        Threshold ($t_{succ}$) & 0.8 & $-$ & $-$ & $-$\\
        $n_{check}$ & 100 & $-$ & $-$ & $-$\\
        $n_{eval}$ & 100 & $-$& $-$ & $-$\\
        Cap ($k$) & 1 & $-$& $-$& $-$\\
        Exploration rate $(\epsilon)$ & 1.0 & $-$ & 1.0 & $-$ \\
        Minimum exploration rate & 0.5  & $-$ & 0.05 & $-$ \\
        Exploration decay & 0.999  & $-$ & $-$ & $-$ \\
        Exploration fraction & $-$  & $-$ & 1.0 & $-$ \\
        Learning rate $(\alpha)$ & 0.05  & 7e-4 & 1e-4 & 3e-4\\ 
        Discount factor $(\gamma)$ & 0.95 & 0.95 & 0.95 & 0.95 \\
        Number of episodes & 5000 & 5000 & 5000 & 5000 \\
        Maximum episode length & 100 & 100 & 100 & 100\\
        \hline
    \end{tabular}
    \caption{Parameters used in Water World.}
    \label{tab:tab4}
\end{table}

\section{Algorithmic Details}
\subsection{Finding Unstable States}
Starting with an initial coarse abstraction, \alg{} proceeds the learning phase by implementing a vanilla Q-learning routine over abstract states to learn an abstract policy $\bar{\pi}(\bar{s})$. Since this coarse initial abstraction is likely to be too coarse, \alg{} checks a refinement condition every $e_{check}$ episodes. This refinement condition evaluates to true if the recent succes rate of the learning agent is below some threshold $t_{succ}$. When the refinement condition evaluates to true, it implies that the current abstraction is not effective and requires further refinement. However, \alg{} doesn't know which abstract states are too coarse. Therefore, \alg{} starts the evaluation phase for $n_{eval}$ episodes to collect some samples of Q-values while the RL agent is interacting with the environment with a fixed policy. Thus, the evaluation phase is simply a Q-learning routine over abstract states with a fixed policy. Since the policy is fixed in this phase, high variation in Q-values for the same state-action pair ($\bar{s}, a$) implies an instability in the abstract state. When the evaluation phase is terminated, all samples are stored in $\Gamma$ from which the refinement phase can be executed. In the refinement phase, $\texttt{UnstableState}(\Gamma)$ takes the collected samples $\Gamma$ and returns the top $k$ unstable states. To this end, $\texttt{UnstableState}(\Gamma)$ calculates the normalized standard deviation of Q-values for all samples of ($\bar{s},a$). In other words, there should be multiple samples in $\Gamma$ for the same abstract state and action pair. When the normalized standard deviation is calculated for all these samples, $\texttt{UnstableState}(\Gamma)$ clusters the abstract states (that have been sampled in $\Gamma$) into stable and unstable states using k-means clustering technique. Finally, $\texttt{UnstableState}(\Gamma)$ returns the top $k$ unstable states from the unstable cluster. 

\subsection{Finding unstable state variables}
When $\texttt{UnstableState}(\Gamma)$ finds the top $k$ unstable states from the samples collected throughout the evaluation phase, the unstable states can be split into $f$ new states w.r.t a state variable $i$ following the definition of f-split refinement in Definition. 2. Then the question is: what state variable should \alg{} blame for the observed instability in an unstable state? 

\textbf{Deliberative Approach}. We know that \alg{} learns an abstract policy $\bar{\pi}$ over abstract states so it maintains and updates the Q-table for abstract states to find the optimal abstract policy. However, \alg{} also maintains and updates the Q-table for concrete states. This concrete Q-table can be further used for various application such as finding contributing state variables for an unstable state. The notion of deliberative refinement is to refine an unstable state w.r.t a state variable that results in the most consistent new abstract states. Basically, splitting an abstract state over a state variable results in $f$ new abstract states. Now, for each newly created abstract state, \alg{} goes through the underlying concrete states and calculates the normalized standard deviation of the Q-values. Intuitively, if all concrete states under any of the newly created abstract states have Q-values with small standard deviation for the same action $a$, then splitting over that state variable would be the near-optimal refinement and can potentially decrease/resolve the instability in the abstract MDP. \alg{} repeats this process for all state variables and chooses the one that minimizes the standard deviation of the underlying Q-values on the concrete level. 

\textbf{Aggressive Approach}. In this approach, \alg{} splits an unstable state w.r.t all state variables. In other word, \alg{} blames all state variables for the instability in an abstract state. This method can be employed to avoid keeping the track of the concrete Q-table. The aggressive \alg{} is essentially effective where the reward is frequent with high variation in an environment.

\begin{figure}[h]
     \centering
         \centering
         \includegraphics[width=0.6\columnwidth]{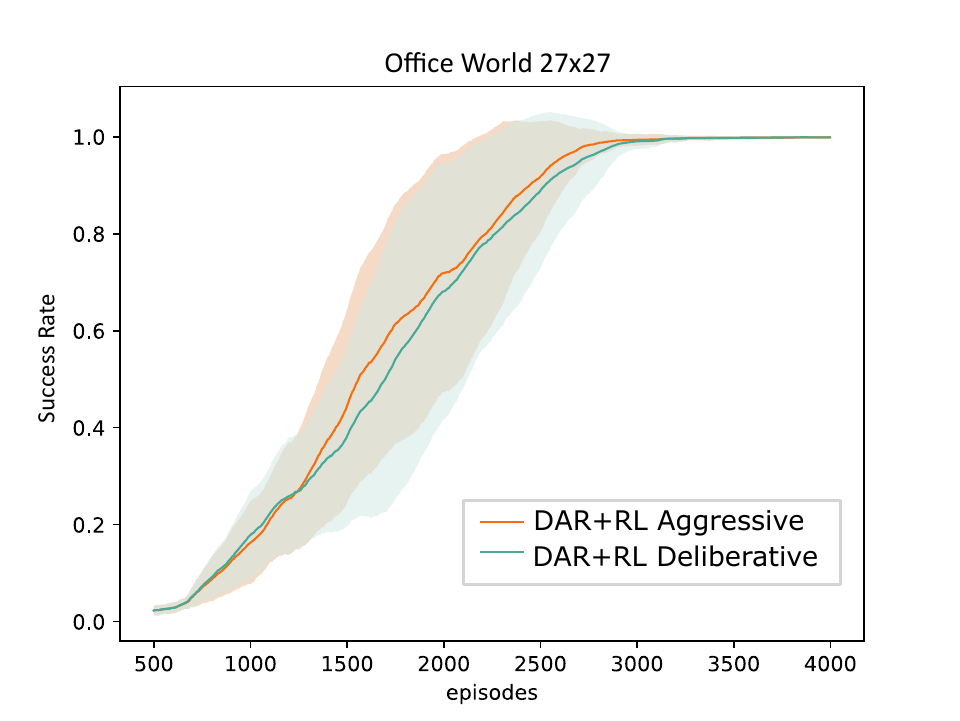}
        \caption{\small Comparing aggressive and deliberative variants of DAR+RL. We ran both variants 10 times and the shadows show the standard deviation of the measurements.
        }
        \label{fig:variants}
\end{figure}

We implemented the aggressive and deliberative \alg{} for blaming a state variable to do the refinement of an unstable state. We analyzed the performance of both variants of DAR+RL in Office World and demonstrated that the DAR+RL performs robustly regardless of the choice of this component, as shown in \ref{fig:variants}.

\section{Domain Descriptions}

\textbf{Office World.}
We consider a 36$\times$36 office world scenario with walls and four rooms A, B, C, and D. The task for the agent is to collect coffee and mail and deliver them to the office. The agent can execute any action from East, West, North, and South. On applying any action, the agent executes the action successfully with probability 0.8 and may slip to one of the two adjacent cells with the probability of 0.1 each. The agent receives a reward of 1000 on completing the task successfully and 0 otherwise. 

\textbf{Wumpus World.} 
We consider a 64$\times$64 Wumpus world with obstacles and pits. The task for the agent is to reach the south-east corner location from the north-west corner location in the grid while avoiding pits. The four actions and the stochastic probabilities are same as in the office world. If the agent's movement is obstructed due to an obstacle, it falls back to its location and receives a reward of -2. The agent receives -1 reward on every step and the episode ends as soon as it enters a pit, receiving a negative reward of -1000. On reaching the correct destination location, it receives a positive reward of 500. 

\textbf{Taxi World.}
We consider a 30$\times$30 taxi world scenario in which there are four pick-up and drop-off locations, one in each corner of the grid. The taxi agent starts at a random cell in the grid. The task for the taxi is to pick up a passenger from its pick-up location and deliver at its destination drop-off location, both selected randomly. It can execute actions: East, West, North, South, Pick-up, and Drop-off. Each move action has stochastic probabilities similar to Office world. It obtains a reward of -1 on applying a move action and -100 on illegal pick-up and drop-off actions. Upon dropping the passenger at the correct destination, it receives a positive reward of 500.  

\textbf{Water World.} We consider a 300$\times$300 two dimensional box with one green ball, one red ball, and one agent represented by a black ball. Each ball moves in one direction with constant speed and bounces back upon hitting the edges. The agent has control over its velocity via taking a move action in one of the east, west, north, and south directions. The task for the agent is to collide with the moving green ball while avoiding the red ball. The episode terminates when the agent collides with a ball. The agent receives a reward of 1000 and -1000 on colliding with the green ball and the red ball respectively. 

\end{document}

%% file: math_commands.tex
\usepackage{amsmath,amsfonts,bm}

\def\eqref#1{equation~\ref{#1}}

\def\1{\bm{1}}

\DeclareMathAlphabet{\mathsfit}{\encodingdefault}{\sfdefault}{m}{sl}
\SetMathAlphabet{\mathsfit}{bold}{\encodingdefault}{\sfdefault}{bx}{n}

